# MetaHarm: Harmful YouTube Video Dataset Annotated by Domain Experts, GPT-4-Turbo, and Crowdworkers

**Wonjeong Jo[1], Magdalena Wojcieszak[1,2]**

[1]Department of Communication, University of California, Davis
[2]Center for Excellence in Social Sciences, University of Warsaw
{wjo, mwojcieszak}@ucdavis.edu

## Abstract

Short video platforms, such as YouTube, Instagram, or TikTok, are used by billions of users. These platforms expose users to harmful content, ranging from clickbait or physical harms to hate or misinformation. Yet, we lack a comprehensive understanding and measurement of online harm on short video platforms. Toward this end, we present two large-scale datasets of multi-modal and multi-categorical online harm: (1) 60,906 systematically selected potentially harmful YouTube videos and (2) 19,422 videos annotated by three labeling actors: trained domain experts, GPT-4-Turbo (using 14 image frames, 1 thumbnail, and text metadata), and crowdworkers (Amazon Mechanical Turk master workers). The annotated dataset includes both (a) binary classification (harmful vs. harmless) and (b) multi-label categorizations of six harm categories: Information, Hate and harassment, Addictive, Clickbait, Sexual, and Physical harms. Furthermore, the annotated dataset provides (1) ground truth data with videos annotated consistently across (a) all three actors and (b) the majority of the labeling actors, and (2) three data subsets labeled by individual actors. These datasets are expected to facilitate future work on online harm, aid in (multimodal) classification efforts, and advance the identification and potential mitigation of harmful content on video platforms.

**Datasets** — https://zenodo.org/records/14647452

## Introduction

There are serious concerns that social media platforms direct users to harmful content. A user interested in herbology may encounter misinformation and a sad adolescent may see depressive content. 66% of adults report encountering harmful content online (Turing Institute 2023), audits indicate that self-harm and suicide posts on Instagram range from 9% to 66% (Picardo et al. 2020), and others find that YouTube often promotes drugs, alcohol, and bullying (Hartas 2021; Hattingh 2021). Accordingly, 90% of US adults state that social media has harmful effects (Statista 2023), and researchers find that social media use leads to mental health issues, eating disorders, or self-harm (Haidt and Twenge 2023; Hartas 2021).

Although platforms regulate harmful content, there are no consistent harm detection criteria across platforms (Arora et al. 2023) and – in fact – some platforms move away from harm moderation altogether (e.g., Meta stopping fact-checking). Further, harm is measured differently across studies (Banko, MacKeen, and Ray 2020; Scheuerman et al. 2020; Shelby et al. 2022) and harm identification approaches are best suited for text-based messages, overlooking the increasingly popular (short) video platforms, such as YouTube, Instagram, or TikTok.

Despite the fact that exposure to online harm has far-reaching implications, we lack a comprehensive understanding, labeling, and datasets of harmful content. The existing datasets contain online spaces (e.g., YouTube channels or web domains) deemed harmful (e.g., Ledwich and Zaitsev 2019, Ribeiro et al. 2020; Allcott and Gentzkow 2017) or compile content (e.g., Liaw et al. 2023; Piot, Martín-Rodilla, and Parapar 2024) that falls under one harm category (e.g., misinformation, misogyny). Although useful, these datasets are limited in advancing our understanding and measurement of online harm. After all, not all content from an online space considered harmful *is* harmful (e.g., not all articles on a fake news site are misinformative) and datasets that focus on one individual harm overlook its multi-categorical nature (e.g. misinformation promoting hate). As importantly, existing resources are primarily text data, not multimodal content that combines image, audio, and text.

This paper offers two systematically curated datasets: (1) 60,906 *potentially* harmful YouTube videos identified using three principled approaches (keyword-based, channel-based, external dataset integration) and (2) 19,422 videos randomly selected from the larger dataset, annotated by trained domain experts, GPT-4-Turbo, and crowdworkers. This annotated dataset includes both (a) binary classification (harmful vs. harmless) and (b) multi-label categorizations of six harm categories: Information, Hate and harassment, Addictive, Clickbait, Sexual, and Physical harms. Furthermore, the annotated dataset provides (1) ground truth data, i.e., videos annotated consistently across (a) all three actors[1] and (b) the majority of the labeling actors, and (2) three data subsets annotated by individual actors separately. For harm classification, we used a curated taxonomy for online harm developed by Jo, Wesołowska, and Wojcieszak (2024).

These comprehensive datasets offer numerous research opportunities. First, they can set a framework for defining, operationalizing, and assessing harm across platforms and projects, which can ultimately facilitate harm detection and moderation. Second, the datasets contain multi-modal and multi-label content, which not only aids in future multi-modal classification efforts but also captures the increasingly prevalent video-based online contexts in which harm is consumed. Third, the sub-datasets with only human or machine annotation can be useful for various research purposes (e.g., LLM-based experimentation). Fourth, the annotated dataset contains content that falls under six harm categories, synthesized from platform community standards and past literature. This goes beyond the fragmented operationalizations and offers more complete insight into the multi-categorical nature of harm and more precise future estimates on exactly how much harmful content there is on platforms. To the best of our knowledge, the labeled dataset is the first large, video-based, and cross-labeled harmful dataset encompassing numerous harm categories.

Below, we describe data collection methods, the labeling process, descriptive statistics, and information on the agreement between the three labeling actors.

## Data Collection

We first had to define what constitutes a harmful video, as the term "harm" is broad and multifaceted. We adopted a taxonomy for online harm curated by Jo, Wesołowska, and Wojcieszak (2024), which synthesizes academic studies and platform guidelines from YouTube, Meta, and TikTok. They categorize harm into: Informational, Hate and harassment, Clickbait harms, Addictive harms, Sexual harms, and Physical harms. Table 1 shows the categories and their corresponding subcategories.

| Harm categories | Subcategories |
|---|---|
| Information harms | Fake news, Conspiracy theories, Unverified medical treatments, Unproven scientific myths |
| Hate and harassment harms | Insults and obscenities, Identity attacks or misrepresentation, Hate speech based on gender, race, ethnicity, age, religion, political ideology, disability, or sexual orientation |
| Addictive harms | Online gameplay, Drug/smoking/alcohol promotion, Gambling-play videos |
| Clickbait harms | Clickbaitive titles, Get-rich-quick schemes or fishing financial hacks, Gossip promotion |
| Sexual harms | Erotic scenes or images, Depictions of sexual acts and nudity, Sexual abuse |
| Physical harms | Self-injury, suicide, Eating disorder promotion, Dangerous challenges and pranks |

Table 1. Taxonomy for online harm by Jo, Wesołowska, and Wojcieszak (2024)

We initiated the construction of the first dataset by compiling 60,906 *potentially* harmful videos that had not yet been annotated for harm. The data collection followed a three-step process outlined below: (1) keyword-based, (2) channel-based, and (3) external dataset integration.

### Keyword-Based Approach

To collect videos relevant to each harm category, we first used 169 keywords in the YouTube search engine. These were developed by integrating keywords from platform community guidelines and from past work (Bouma-Sims and Reaves 2021; Chancellor, Kalantidis, and Pater 2017; Hussein, Juneja, and Mitra 2020; Kriegel et al. 2021; Paek et al. 2014; Scherr et al. 2020).

We utilized YouTube Data API v3[2] in a Python environment to collect text-formatted metadata (i.e., video title, channel name, publication date, view count, and description). Transcripts were extracted using the YouTubeTranscriptApi[3] and linked to each video via its URL. The keywords were entered into the API, and search results were filtered using both recency and relevance filters in a 7:3 ratio to timely collect videos that may be quickly removed by the platform. While relevance filters are effective for capturing widely viewed videos, the recency filter yields the latest uploads. The recency filter was particularly useful for subcategories under physical harms, such as self-injury and suicide, where relevance filters prioritize prevention and treatment videos. We collected the search outcomes from October 2023 to January 2024 using the English search queries and phrases listed in Table 2.

| Harm categories | Search queries and phrases |
|---|---|
| Information harms | government lies about the dark truth, cover -ups and conspiracies revealed, chemtrails, weather manipulation theories, flat Earth conspiracy evidence, Moon landing hox arguments, 9/11 truth movement evidence, Area 51 secrets and conspiracies, world order and global governance, climate change |

[1] In this manuscript, the term "actor" refers to the annotating entities: domain experts, GPT-4-Turbo, and crowdworkers.

[2] https://developers.google.com/youtube/v3/getting-started

[3] https://pypi.org/project/youtube-transcript-api/

| | |
|---|---|
| | denial, vaccine injuries, suppressed medical breakthrough, Proven miracle cure, alternative medicine remedies, herbal remedies for, big pharma scandals, miracle pill that cures cancer, the dark side about, exposing lies, debunking mainstream narratives, top-secret government experiments, hidden agendas, untold stories, vaccines and autism, vaccine dangers, mushroom cures HPV, natural cures for HPV |
| Hate and harassment harms | fuck lgbt, fuck gay, fuck black, fuck Asian, fuck transgender, lgbt hate, gay hate, Asian hate, gay-sex, disability sex, funny Asian accents, transgender shocking reality, Asian look alike, trans mental illness, feminism ruining society, manhaters, gay flamboyant, transgender therapy, deaf compilation, blind people compilation |
| Addictive harms | Sudden Attack gameplay, Battle gameplay, gamer paradise, gamer showdown, streamer's choice, game highlight, casino hits, casino play, casino trip, win Roulette, online gambling highlight, big wins, super slot, gambling, smoking, puffing, toking, drunk, day drinking, cigarette, cigarette scene, on weed, weed trip, smoking for the first time, cannabis, marijuana |
| Clickbait harms | get rich quick, easy money, instant cash, top 10, you won't believe, unbelievable, must see, caught on camera world-ending, life-changing, guaranteed results, free giveaway, one-time opportunity, 100% safe, 100% free, breaking the Internet, endless cast flow, shocking celebrity transformation, hottest gossip, the scandals, free trial, make thousands per day, double your investment, limited time offer, exclusive access, secret method, free account |
| Sexual harms | sex movie, sex video, woman boob, 18+ adult movie, sexy hot girl, upskirt, porno movie, porno, provocative woman, public masturbation, seductive scene, sexy hot girl, erotic movie, uncensored body cam, nude body, erotic scene, upskirt, 18+, hot woman, sexy woman, adults only |
| Physical harms | self-harm, deliberate self-harm, non-suicidal self-injury, will suicide, #suicide, self-cutting, suicide, selfcutting, self harm, risky challenge, thinspo, proana, promia, eating disorder, thinspiration, depression, skinny body, thigh gap, thin waist, help harm, suicidal, depression risk, car accident, ufc, gun shot, shooting, firearm, kill scene, crime scene, dead body, deadlist challenge, risky challenge, abusive TikTok, junk food challenge, fight blood, this video got deleted, self harm report, tough trial, giant challenge, dangerous experiments, extreme dare, The shocking murders on Camera, animal cruelty, animal abuse TikTok, egg crack challenge, blackout challenge, tide pod challenge, animal abuse TikTok, hanging dog, dog death, cat death, human test, shocking murders |

Table 2. Search Queries for Keyword-Based Approach

## Channel-Based Approach

In addition, we identified *potentially* harmful YouTube channels and scraped all videos therein. An undergraduate research assistant created a list of 49 channels categorized by harm type (see Table 3). The channels were selected based on information from online communities (e.g., Reddit) and included channels recognized for spreading harmful content. This list includes a relatively small number of channels from information harms and physical harms, so we supplemented the former with videos from external datasets (detailed below) and by generating the extensive set of keywords for physical harm (see Table 2). We also relied on external datasets for hate and harassment harm.

We used Python libraries, including `requests` and `Selenium`. Using `Selenium`'s WebDriver, we automated browsing to navigate each channel's page and scrape video URLs by extracting the HTML elements containing video links. After collecting the video URLs, we retrieved metadata and transcript for each video link, using the YouTube Data API v3 and YouTubeTranscriptApi.

| Harm categories | Subcategories |
|---|---|
| Information harms | @VineMontanaTV, @VoyagerSpace, @X22Report, @The Next News Network |
| Addictive harms | @cewpins, @MrTheManWeeD, @WelcometoTheGrowTent, @CustomGrow420, @xCodeh, @StrainCentral, @urbanremo, @ROSHTEINBIGWINS, @OnlineGamblingHighlights, @HiiiKey, @MrMikeSlots, @WatchGamesTV, @damianluck925, @MC.Nguyena9, @XEPOJmx, @MMOHut |
| Clickbait harms | @lucasmorocho09, @CelebritiesTV, @loganpaulvlogs, @BusyFunda, @growify, @mwmedia4421, @SmartMoneyTactics, @JohnCrestani, @CurrencyCounts, @GameJacker, @MeetKevin, @BRIGHTSIDEOFFICIAL, @5MinuteCraftsPLAY, @IULITM, @TroomTroom, @JazeCinema, @FactsVerse |
| Sexual harms | @BloomingDerek-fu6jf, @MyTinySecretsTV, @LoveIsLife, @lamAj, @OfficialSecretDiary, @Raanjhna, @HOTSEX24, @tubetv2445, @ViaViolaChannel |
| Physical harms | @superhumman, @AustralianSparkle, @light_as_a_feather1497 |

Table 3. YouTube Channels for Channel-Based Approach

### External Dataset Integration

We incorporated videos from external datasets for Information harms and Hate and harassment harms. For the former, we integrated thousands of conspiracy theory videos from an external dataset (YouNICon; Liaw et al. 2023) and COVID-related misinformation videos (Knuutila et al. 2021). For Hate and harassment, we included channels associated with anti-feminist, white identitarian (Ledwich and Zaitsev 2019; Mamié, Ribeiro, and West 2021), and Intellectual Dark Web (Ribeiro et al. 2020). We scraped sample videos from these channels and retrieved metadata and transcripts.

### Image Frames and Thumbnails Retrieval

Through this approach, we identified 60,906 unique *potentially* harmful videos, 10,000 per category. Next, out of this large pool, we sampled 19,422 videos across the six harm categories for annotation. 15 image frames and 1 thumbnail image were extracted per each sampled video. To extract frames, we randomly selected 15 frame positions within each video from the first to the second-to-last frame. We utilized the `yt_dlp` Python package to download the videos and the `opencv` library to determine the total frame count. In a loop running 15 times, we generated random frame numbers within the specified range, set the video to the corresponding frame, and captured the image. The thumbnail images were retrieved directly from YouTube using the `yt_dlp` and `pytube` libraries. In this regard, the 19,422 sampled videos were linked to text-formatted metadata (e.g., title, description, channel name, transcript) as well as image data, including 15 image frames and a thumbnail.

## Annotation Process

We annotated the 19,422 videos by (a) determining if they were harmful or harmless and (b) assigning *one or more* specific harm categories. For annotation, we relied on trained domain experts, non-trained crowdworkers, and a Large Language Model (LLM), GPT-4-Turbo. In past scholarships, expert annotators are often served as gold standard or as platform moderators (Kim, Kim, and Jo 2024; Ostyakova et al. 2023) and crowdworkers are frequently used to annotate training data for classifier development (Plank 2022). Recently, LLMs are increasingly employed as an alternative in annotation tasks (Rathje et al. 2024; Ostyakova et al. 2023).

### Domain Expert Labels

Ten students majoring in communication and digital media served as domain experts. They underwent extensive training to familiarize themselves with the harm taxonomy and coding instructions before and during the annotation process. The instructions incorporated the harm taxonomy, subcategory examples, and sample videos for each category. The experts classified each video into one or more categories or labeled it as harmless or unavailable (for removed videos). For videos longer than 5 minutes, the experts were instructed to skim the content by adjusting the mouse pointer or playback speed.

Each video was labeled by a single expert. However, 11.4% of the videos (n = 2,225) were annotated by more than two experts. This subset was used for calculating intercoder reliability (ICR), with final labels determined through majority agreement. ICR was 0.88 using Holsti's index (1969) for the binary classification task and 0.76 using Cohen's Kappa for the multi-label classification. We also calculated the ICR between the experts for each harm category. The percentage agreement based on Holsti's index was highest for Addictive harms (87.36%) and lowest for Hate and harassment (58.84%). Further details of the ICR calculation are provided in Jo, Wesołowska, and Wojcieszak (2024). The experts conducted the classification over four months, and the entire procedure was approved by the IRB of the authors' institution.

### GPT-4-Turbo

GPT-4-Turbo was the second labeling actor. It is OpenAI's Multimodal Large Language Model (MLLM) released in November 2023 (most recent MLLM available at the time). For classification, we fed a thumbnail, image, 14 image frames, and text metadata (title, uploader, description, and transcript) along with the prompt into the GPT API model of `gpt-4-turbo-2024-04-09`. Each image was resized to 768 pixels and converted to base64 format. The transcript was truncated to a maximum of 3,000 words to handle lengthy videos. We set the API model temperature to 0.7, indicating a moderate level of randomness in the outputs for each iteration, and limited the output tokens to a maximum of 25. `OpenAI` Python package was used.

The prompt was designed as a zero-shot learning prompt that comprises video information and the coding instruction (online harm categories) (see Figure 1). Specific classification examples were not provided to avoid constraining the classification reasoning to a narrow set of predefined examples, given the wide range of possible cases in each harm category. In that regard, zero-shot trained prompts have revealed strong performances in annotation tasks (Toney-Wails, Schoeberl, and Dunham 2024; Törnberg 2024; Sekharan and Vuppala 2023).

**[Image frames]** *14 image frames and 1 thumbnail image are provided here*
**[Task assignment]**

Your task is to classify YouTube videos as Harmful or Harmless, incorporating the image frames and metadata of a YouTube video. Please classify this video, making sure to reference the [Coding Instruction].

**[Coding Instruction]**
1. Information Harms (e.g., Fake news, Conspiracy theories, Unverified medical treatments, Unproven scientific myths)
2. Hate and Harassment Harms (e.g., Insults, Obscenities, Identity attacks or misrepresentation, Hate speech based on gender, race, religion, disability, sexual orientation)
3. Addictive Harms (e.g., Gameplay, Drug/smoking/alcohol promotion, Gambling videos or promotion)
4. Clickbait Harms (e.g., Clickbait titles - using exaggerated words, "You won't believe," "10 reasons you don't know," "Best/easiest hacks," Get-rich-quick schemes and Fishing financial hacks, Gossip about celebrities)
5. Sexual Harms (e.g., Erotic scenes or images, Depictions of sexual acts and nudity, sexual abuse)
6. Physical Harms (e.g., Self-injury, Eating disorder promotion, Potentially dangerous challenges and pranks, Violent graphic content)

**[Metadata]** *title, channel information, description, and transcript are provided here*

**[Question]**
Your answer should follow this format:
1) Harmful / Harmless
2) Corresponding one or multiple categories IF HARMFUL else None

Figure 1. Prompt for GPT-4-Turbo Classification

For classification, we generated three API keys for GPT-4-Turbo classification and ran them on each video. We then selected the majority response to control the randomness of LLMs. Since the model is built on probabilistic programming, it can produce varying outputs for the same input across instances. By capturing agreement among multiple responses, we aimed to reach reliable output. For example, if two APIs classified a video as harmful while the third classified it as harmless, or if all three APIs did so, the video was determined as harmful. If two or three API keys refused to answer, or if two provided conflicting responses while one refused, we repeated the prompt with three new API keys to determine the response. 227 videos (1.7%) remained unresolved due to no majority answer or consistent refusals even after this iteration.

ICR among the three GPT APIs was assessed on 30% of the dataset. Krippendorff's alpha measuring overall agreement was 0.78.

### Crowdworkers

Lastly, we relied on Amazon Mechanical Turk (MTurk) for crowdworker labels. We recruited master-level users with a HIT approval rate above 95%. This indicates they consistently deliver high-quality work and that the majority of their submissions have been approved by requesters, thus expecting relatively reliable output from these workers. Only workers aged over 18 and self-reported English speakers were eligible to participate. To increase quality, we set five filter questions at the start of the task (pass rate = 93.89%). Only crowdworkers who passed these questions could proceed.

544 workers were recruited and paid $2 as compensation per task. Their task was to classify 25 videos as harmful, harmless, or unavailable (if removed), and to select one or more harm categories for those labeled as harmful. For videos over five minutes, they were instructed to skim the content as in expert training. The median time taken to complete the task was found to be 18.37 minutes. Similar to the GPT-4-Turbo classification, each video was labeled by three workers, and the majority agreement was selected.

We calculated ICR on a random 30% sample. In line with prior studies pointing out the limited reliability of open-sourced crowdworkers, particularly MTurkers (Chmielewski and Kucker 2020), Krippendorff's alpha among the three crowdworkers was the lowest at 0.21. That said, because final labels were determined based on a majority agreement among at least two workers, individual-level variance was, we believe, partially mitigated in the final annotated dataset.

## Data Structure

We largely offer two datasets: (1) 60,906 unannotated large pool as *potentially* harmful videos and (2) 19,422 annotated videos organized into separate folders for text metadata and image data. Below, we outline the folder structure and data attributes.

### Folder Information

Table 4 presents the overview of the folders.

#### Unannotated Large Pool

This folder contains text-formatted metadata (titles, uploaders, descriptions, transcripts, publication dates) of the 60,906 potentially harmful videos. Note that we do not claim that all included videos are harmful, as not every video related to keywords or channels is necessarily harmful within platform environments. Unlike the annotated videos, these are not linked with any associated images.

#### Ground Truth

This folder contains text metadata of the *harmful* videos identified through overlapping classification results across the annotation actors. We expect these videos to function as ground truth data for harmful YouTube videos.

**Harmful_full_agreement.** This folder contains videos classified as harmful by *all three actors*: domain experts, GPT-4-Turbo, and Mturk workers.

**Harmful_subset_agreement.** This folder contains videos annotated as harmful by *at least two out of the three annotator groups*. For example, it includes videos labeled as harmful by both domain experts and GPT but not by crowdworkers, or any other combination involving agreement among two or more actors.

**Domain Expert**

This folder contains text-formatted metadata of videos annotated as harmful or harmless by domain experts, organized into separate subfolders.

**GPT-4-Turbo**

This folder contains text-formatted metadata of videos classified as harmful or harmless by GPT-4-Turbo. Both the binary labels (harmful vs. harmless) and specific harm categories reflect the majority response of the three API keys.

**Crowdworker**

This folder contains text-formatted metadata of videos annotated as either harmful or harmless by MTurk workers. The resulting binary labels and harm categories reflect the majority determination among the three workers.

**Thumbnails**

This folder provides thumbnails from the classified videos. Each thumbnail is named after the unique video ID. This folder does not distinguish between videos classified as harmful or harmless.

**Image Frames**

Image frames are organized into 19 zip folders. Each zip folder contains subfolders named after the unique video IDs of the annotated videos. Inside each subfolder, there are 15 sequentially numbered image frames (from 0 to 14) extracted from the corresponding video. The image frame folders do not distinguish between videos classified as harmful or non-harmful.

## Exploratory Analysis

We present an exploratory analysis outlining our dataset. First, we compare the distribution of harm categories identified by domain experts, GPT-4-Turbo, and crowdworkers. Next, we provide a descriptive overview based on the videos' metadata. Lastly, we examine the extent to which GPT-4-Turbo and crowdworker labels align with those of domain experts.

### Harm Categories Distribution

**Ground Truth Data**

Out of the 19,422 annotated videos, 5,109 (26.3%) were classified as harmful with full agreement among all three actors, while 14,019 (72.2%) were classified as harmful with agreement by at least two actors with subset agreement. Meanwhile, only 589 (3.0%) videos were classified as harmless with full agreement among all three actors, and 3,980 (20.5%) were classified as harmless with subset agreement by more than two actors (see Figure 2).

| Format | Folder | Subfolder | #Videos (%) |
|---|---|---|---|
| Text | Unannotated Large Pool | | 60,906 |
| | Ground Truth | Harmful_full_agreement | 5,109 (26.3%) |
| | | Harmful_subset_agreement | 14,019 (72.2%) |
| | Domain Experts | Harmful | 15,115 (77.82%) |
| | | Harmless | 3,303 (17.01%) |
| | GPT-4-Turbo | Harmful | 10,495 (54.03%) |
| | | Harmless | 7,818 (40.25%) |
| | Crowdworker | Harmful | 12,668 (65.22%) |
| | | Harmless | 4,390 (22.6%) |
| Images | Thumbnails<br>Image Frames | | 19,422 |

*Note.* The percentages were calculated based on the total number of annotated videos (19,422). Some videos were classified as unavailable because no majority agreement was reached or the videos were removed. As a result, the sum of harmful and harmless video labels for each actor does not total 100%.

Table 4. Dataset Folders Overview

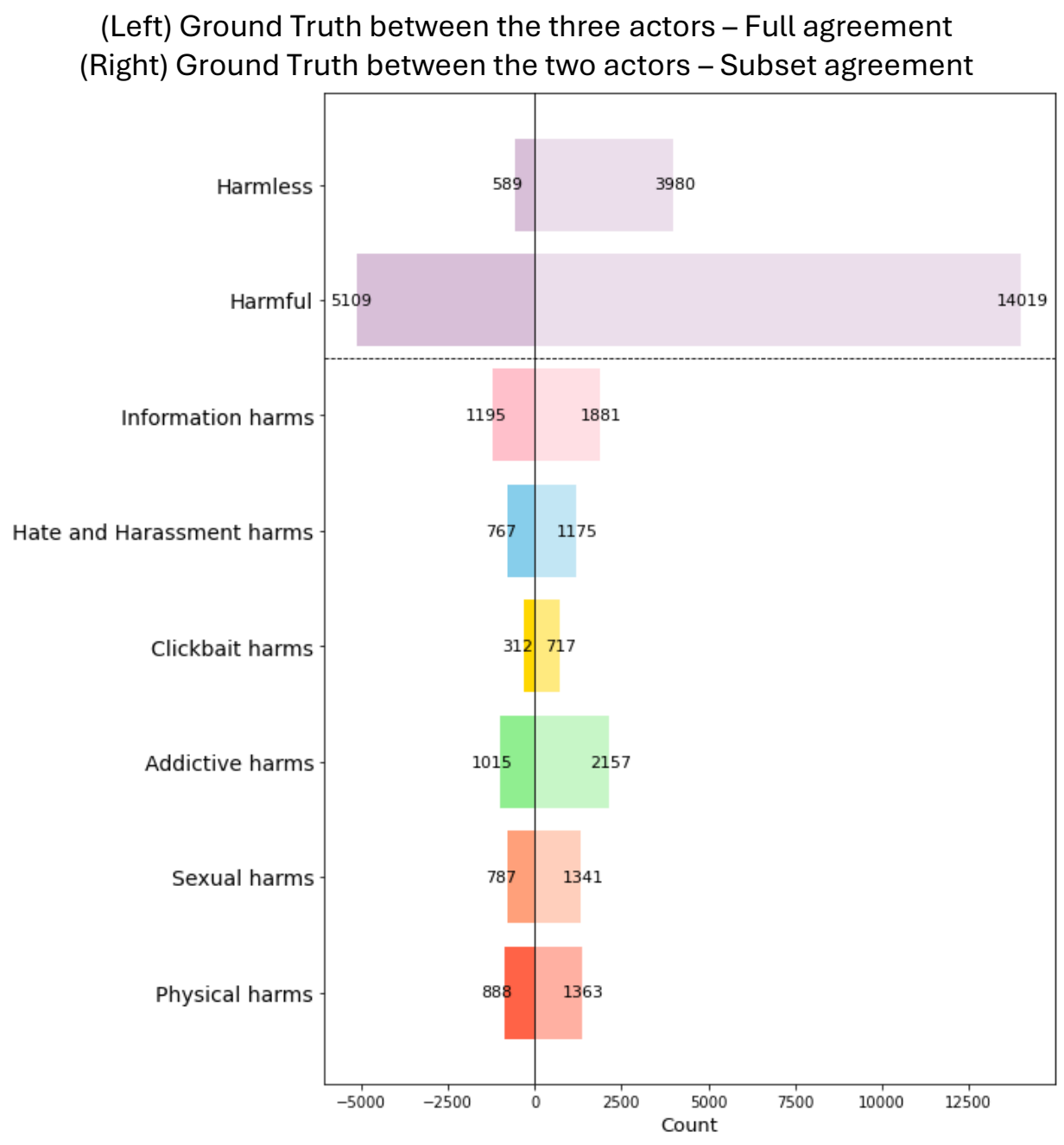

Figure 2. Classification Distribution of Ground Truth Data

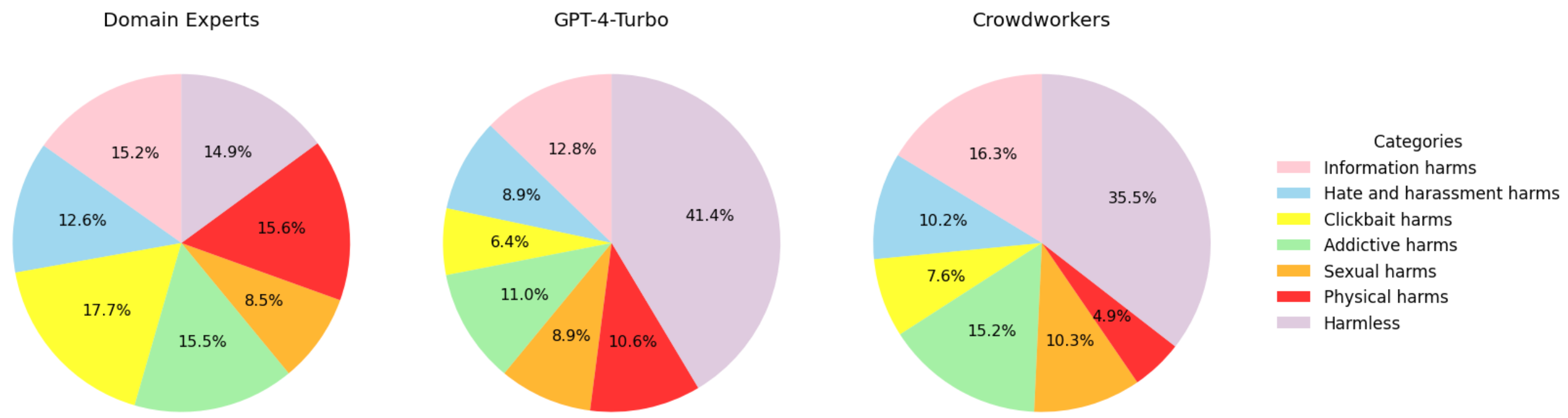

Figure 3. Harm Category Distribution

In Figure 2, the six harm categories under the *harmful* label represent the harms assigned to videos identified as harmful. For full agreement cases, the harm categories were assigned based on agreement among all three actors. For the subset agreement ones, the categories were assigned based on agreement among at least two actors. In both the full and subset agreement cases, the harm categories are evenly distributed with the exception of Clickbait harms. The number of videos labeled under Clickbait is found to be lower compared to the other categories overall.

Even if a video was identified as harmful through majority determination across actors, it was not assigned a ground truth harm category unless there was agreement among the actors on the specific harm category.

### Individual Classification Outcome

Domain experts classified 77.82% of the 19,422 videos as harmful, compared to 54.03% by GPT-4-Turbo and 65.22% by crowdworkers. To harmful videos, each actor assigned one or more harm categories. The proportion of classifications varied across categories for each labeling actor (see Figure 3). For example, domain experts most frequently labeled clickbait harm (17.7%), whereas GPT-4-Turbo and crowdworkers primarily identified information harm (12.8% and 16.3%, respectively) on the same video sets.

## Descriptive Overview

Figure 4 shows the temporal distribution of published dates for the 60,906 *potentially* harmful videos and the annotated 19,422 videos. Approximately half were published after 2022. Of the 19,422 classified videos, 25.8% were uploaded after 2023, and 18.6% were from 2022. The keyword-based approach for data collection done from mid-2023 to early 2024 likely influenced this pattern by capturing recent videos. Additionally, it is possible that YouTube removed explicitly harmful videos published in earlier years, making them inaccessible during data collection.

Figures 5 and 6 show the word count distributions for the descriptions and transcripts of the annotated videos. Both follow a power-law distribution where shorter descriptions and transcripts are most common and longer ones are increasingly rare.

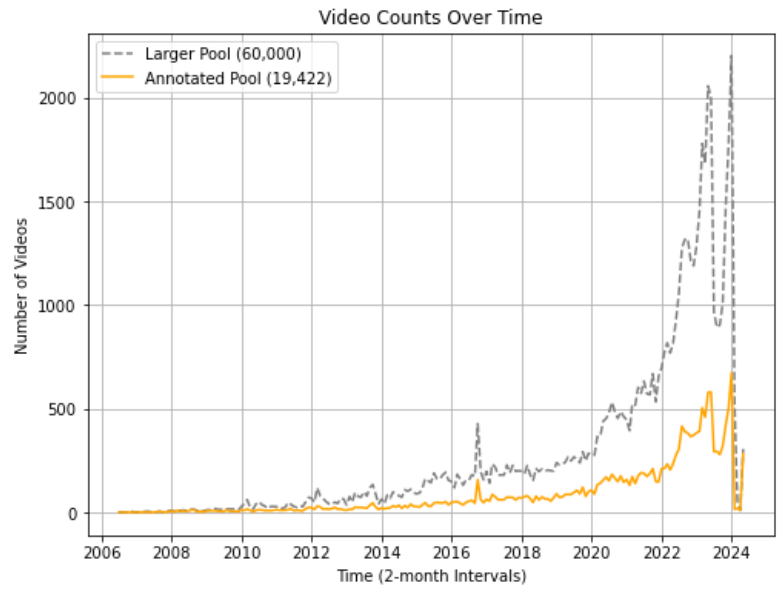

Figure 4. Number of Videos Collected over Time

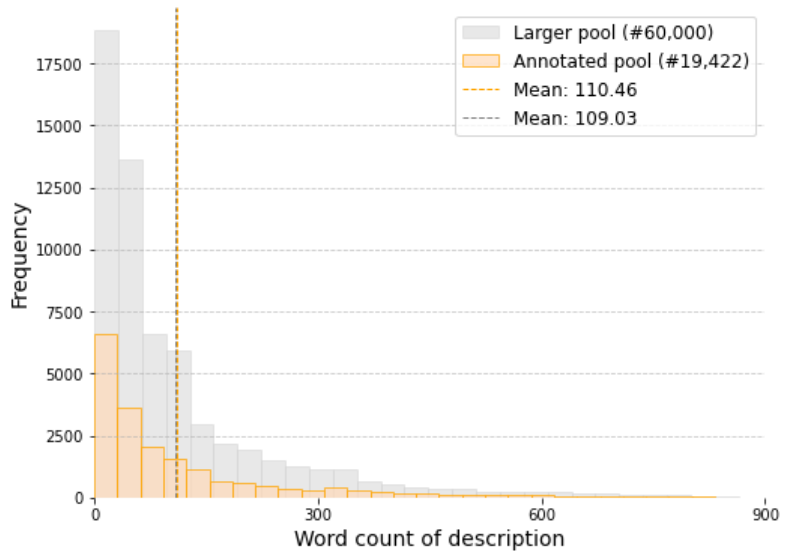

Figure 5. Word Count Distribution of Description

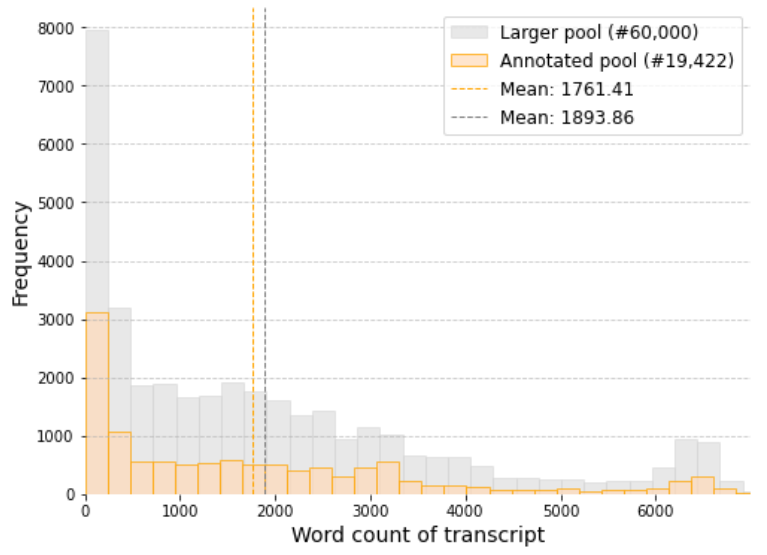

Figure 6. Word Count Distribution of Transcript

## Actors Comparison

We examined the extent to which GPT-4-Turbo and crowdworker labels align with those of domain experts for the binary classification (harmful vs. harmless). We treated the domain expert labels as actual labels due to their extensive training and domain-specific knowledge.

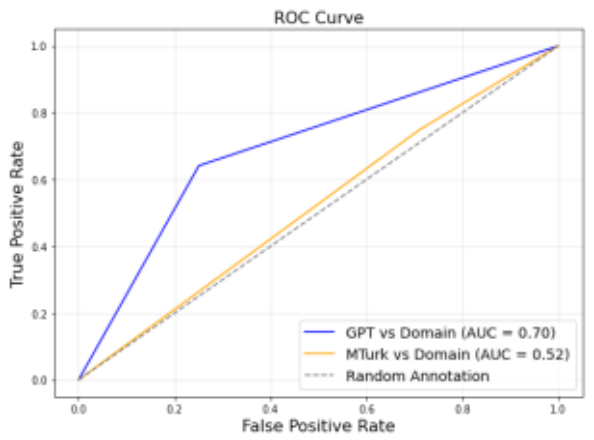

Figure 7. ROC Curve

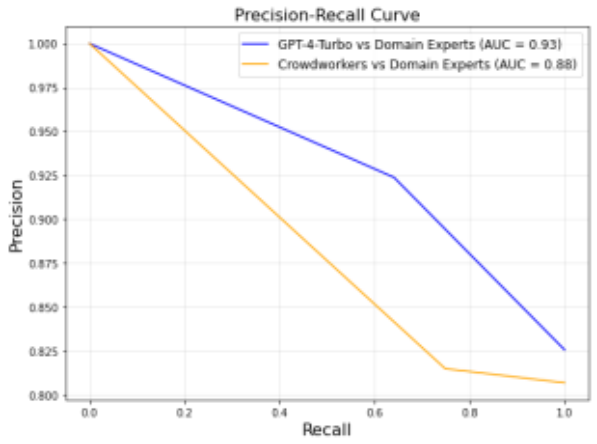

Figure 8. PR Curve

Figure 7 shows a Receiver Operating Characteristic (ROC) curve that represents the relationship between true positive rate and false positive rate across multiple classification thresholds. In classification tasks, an ideal classifier produces a curve that closely follows the top-left corner of the plot, which indicates high sensitivity and low false positive rates. The further a curve is positioned above and to the left of the diagonal line, the better the classifier's performance. In Figure 7, the curve for GPT-4-Turbo is positioned above and to the left of the curve for crowdworkers when compared to the domain expert labels. The Area Under the Curve (AUC), a scalar measure quantifying the ROC curve's performance, was 0.70 for GPT and 0.52 for crowdworkers.

We also report the Precision-Recall (PR) curve (Figure 8) to account for our imbalanced dataset where harmful classes outnumber harmless classes for all annotating actors. In a PR curve, AUC denotes the trade-off between precision and recall, and higher values denote better performance. We see GPT-4-Turbo achieves higher AUC (= 0.93) than crowdworkers (AUC = 0.88) compared to domain expert labels. Putting them together, we conclude that there are more shared labels between GPT-4-Turbo and domain experts than between crowdworkers and domain experts. Further detailed performance evaluations are available in Jo, Wesołowsk, and Wojcieszak (2024).

## Discussion and Contributions

The growth of social media has substantially increased users' exposure to harmful content, one that is sexually explicit, violent, racist or sexist, misinformative or conspiratorial, or promoting financial scams, self-harm, or eating disorders. Despite the prevalence of online harm, its grave effects, and the critical need to minimize it, we lack systematic, multimodal, and multi-categorical datasets of online harms.

To fill this gap, we release two large-scale datasets: (1) 60,906 *potentially* harmful YouTube videos identified using three principled approaches (keyword-based, channel-based, external dataset integration) and (2) 19,422 videos randomly selected from the larger dataset, which were annotated as harmless or harmful and – if the latter – as representing one or more harm categories by three actors: domain experts, LLMs, and crowdworkers. The broad coverage of our datasets and the multi-modal and multi-categorical nature of the annotated dataset contribute to the research community interested in platforms, algorithmic recommendations, online harm, multi-modal messages, and machine learning and artificial intelligence. The large-scale datasets also make the study of numerous research questions possible.

First, categorizing harmful video content into multiple categories is a non-trivial task and – furthermore– what constitutes harm may differ depending on user groups. For that reason, our dataset includes the ground truth based on consensus *and* the majority determination across the three labeling actors, as well as datasets annotated by each individual actor. This may enable research into biases or hallucinations in large language models by comparing them to human annotations and evaluating the impact of different annotation methods on classification outcomes or patterns.

Second, while different studies and datasets available to the scientific community have distinct foci, e.g., specific harms such as hate speech or misinformation, our annotated dataset offers six non-mutually exclusive harm categories applicable to increasingly popular video platforms and adapted from past research and platform community standards. When used to develop multi-modal and multi-categorical harm classifications, as we suggest below, this can facilitate comparative research that estimates the prevalence of distinct harms across platforms (e.g., Is clickbait more prevalent than hate and harassment? Is it more or less prevalent on YouTube versus TikTok?).

Third, our data can be used to train AI harm classifiers for predicting whether content is harmful and what harm category or categories it represents. Our annotated dataset can facilitate classification approaches that are scalable (i.e. require minimal supervision), dynamic (i.e. can adapt to different harm categories at inference time), and robust (i.e. attain high performance even with different harm categories) and – crucially – are well adapted to short-video-based platforms. For instance, scholars could systematically explore whether incorporating visual information alongside text data improves model performance or experiment with adjusting the number of input image frames. Ultimately, the annotated dataset, paired with advancements in machine learning and artificial intelligence, may support the development of tools, models, or software for identifying harmful videos online by expanding traditional methods of harm classification.

Last but not least, the annotated dataset, with video metadata and robust labeling, may have downstream implications for policies surrounding content moderation. We hope it can be used to decrease recommendations and exposure to harm and promote engagement with non-harmful content on social media platforms.

In sum, this dataset advances research on detecting and minimizing online harm. Because "harmful content" is an expansive term understood and measured differently across platforms and by different researchers, the annotated dataset offers the much-needed integration of past work and sets the stage for more comparable research across sciences.

## Limitations

Our data collection focused on YouTube because it is the most popular platform, used by 95% of teens (Vogels and Gelles-Watnick 2023) and 81% of the American population (Auxier and Anderson 2021), and one that is criticized for facilitating exposure to misinformation, stereotypes, and distressing content (Hilbert et al. 2023; Srba et al. 2023). This dataset, therefore, does not capture harm on other platforms.

We acknowledge that how we collected videos may not fully capture the breadth of existing harmful content. Although we aimed to be as comprehensive as possible through a three-step approach considering search queries, channels, and external datasets, we may have missed videos that were already removed by the platform under specific keywords. The annotated dataset only includes videos that were still retrievable from YouTube and we did not have access to the removed videos. To the extent that some content was removed precisely because it was harmful, our annotated dataset does not contain videos that sent the clearest signal about a certain harm. Also, our approaches may not reflect the full spectrum of relevant content circulated via personal algorithms.

When it comes to labeling, the performance of LLMs varies depending on prompt design (Barrie, Palaiologou, and Törnberg 2024; Yu et al. 2023), text length (Heseltine and von Hohenberg 2023; Kim, Kim, and Jo 2024), and task type and context (Kristensen-McLachlan et al. 2023). Additionally, as a closed-source model, LLMs have limited replicability, with their training processes remaining a black box (Kristensen-McLachlan et al. 2023). Future work should examine whether the outcomes of our GPT-based annotation would vary with different video data or image frame selection methods. Notably, our LLM-based classification relies solely on Open AI's GPT-4-Turbo APIs. Incorporating other emerging models, such as Meta's Llama, Google's Gemini, or Anthropic's Claude, may yield more convincing results, which could not be addressed in this manuscript.

## Ethical Statement

This study was approved by the IRB of the authors' institution. All human participants provided informed consent. The consent form outlined potential risks associated with viewing harmful videos and granted participants the right to withdraw from the annotation process at any time. In particular, for the domain expert annotation process, we maintained frequent communication and regularly checked in on their emotional state, accommodations, workload, and scheduling of breaks.

There may be privacy concerns when using the GPT API to process video information. The authors confirmed that OpenAI does not use input or output data from the API for training purposes or share such data with other users by default.[4] Additionally, we kept the used API keys confidential during the analysis and discarded them afterward.

We also caution against the use of videos classified as *harmless* in our dataset, as all videos were sourced from a large pool of potentially harmful videos. Given that the harmful YouTube videos may pose risks to viewers, we encourage readers to use the dataset for socially beneficial purposes. We hope our multimodal dataset encompassing comprehensive harm categories will serve as training data for a wide range of harm mitigation studies. Any misuse of the dataset, such as reproducing abusive and malicious content for personal or commercial purposes, is not permitted.

## FAIR Data Release

Below, we clarify how our dataset release aligns with the FAIR principles (Wilkinson et al. 2016).

- **Findable**: The data is found at Zenodo data repository (https://zenodo.org/records/14647452) with a digital object identifier (DOI): 10.5281/zenodo.14647452
- **Accessible:** The Zenodo repository is publicly discoverable and licensed under CC BY 4.0. Access to the full text data is restricted due to concerns about potential misuse but can be freely requested directly through the platform. Image data are readily available via cloud storage hosted and managed by UC Davis on a Box server.
- **Interoperable**: Text data are formatted as Excel files, image frames are stored in PNG format, and thumbnails in JPG format.
- **Reusable**: Even if original YouTube links become inactive (e.g., due to video removal), the dataset remains reusable because each link is associated with preserved metadata and image data.

## Acknowledgments

The authors are sincerely grateful to Hong An Tran, Dior Tran, Ashtyn Phan, Ary Christine Quintana, Jacob Feinstein, Chushan Huang, Olivia He, Janghee Oh, Katharine Owen, Kristen Ya, Betty Su, Kaitlyn Tan, and CJ Verrengia for their wonderful and thoughtful research assistance.

[4] https://help.openai.com/en/articles/5722486-how-your-data-is-used-to-improve-model-performance

Also, the authors gratefully acknowledge the support of the Excellence Initiative - Research University, University of Warsaw (Priority Research Area V) (Magdalena Wojcieszak -- PI). The authors are also grateful for the support of the Hyundai CMK Scholarship Foundation. Any opinions, findings, and conclusions or recommendations expressed in this material are those of the authors and do not reflect the views of the funders.

## Paper Checklist

1. For most authors...
   (a) Would answering this research question advance science without violating social contracts, such as violating privacy norms, perpetuating unfair profiling, exacerbating the socio-economic divide, or implying disrespect to societies or cultures? Yes
   (b) Do your main claims in the abstract and introduction accurately reflect the paper's contributions and scope? Yes
   (c) Do you clarify how the proposed methodological approach is appropriate for the claims made? Yes
   (d) Do you clarify what are possible artifacts in the data used, given population-specific distributions? Yes
   (e) Did you describe the limitations of your work? Yes
   (f) Did you discuss any potential negative societal impacts of your work? Yes
   (g) Did you discuss any potential misuse of your work? Yes
   (h) Did you describe steps taken to prevent or mitigate potential negative outcomes of the research, such as data and model documentation, data anonymization, re-sponsible release, access control, and the reproducibility of findings? Yes, see the FAIR Data Release and Ethical Statement sections.

(i) Have you read the ethics review guidelines and ensured that your paper conforms to them? Yes

2. Additionally, if your study involves hypotheses testing...

(a) Did you clearly state the assumptions underlying all theoretical results? NA

(b) Have you provided justifications for all theoretical results? NA

(c) Did you discuss competing hypotheses or theories that might challenge or complement your theoretical results? NA

(d) Have you considered alternative mechanisms or explanations that might account for the same outcomes observed in your study? NA

(e) Did you address potential biases or limitations in your theoretical framework? NA

(f) Have you related your theoretical results to the existing literature in social science? NA

(g) Did you discuss the implications of your theoretical results for policy, practice, or further research in the social science domain? NA

3. Additionally, if you are including theoretical proofs...

(a) Did you state the full set of assumptions of all theoretical results? NA

(b) Did you include complete proofs of all theoretical results? NA

4. Additionally, if you ran machine learning experiments...

(a) Did you include the code, data, and instructions needed to reproduce the main experimental results (either in the supplemental material or as a URL)? NA

(b) Did you specify all the training details (e.g., data splits, hyperparameters, how they were chosen)? NA

(c) Did you report error bars (e.g., with respect to the random seed after running experiments multiple times)? NA

(d) Did you include the total amount of compute and the type of resources used (e.g., type of GPUs, internal cluster, or cloud provider)? NA

(e) Do you justify how the proposed evaluation is sufficient and appropriate to the claims made? NA

(f) Do you discuss what is "the cost" of misclassification and fault (in)tolerance? NA

5. Additionally, if you are using existing assets (e.g., code, data, models) or curating/releasing new assets, **without compromising anonymity...**

(a) If your work uses existing assets, did you cite the creators? Yes

(b) Did you mention the license of the assets? NA

(c) Did you include any new assets in the supplemental material or as a URL? Yes

(d) Did you discuss whether and how consent was obtained from people whose data you're using/curating? Yes

(e) Did you discuss whether the data you are using/curating contains personally identifiable information or offensive content? Yes

(f) If you are curating or releasing new datasets, did you discuss how you intend to make your datasets FAIR (see FORCE11 (2020))? Yes, see the FAIR Data Release section.

(g) If you are curating or releasing new datasets, did you create a Datasheet for the Dataset (see Gebru et al. (2021))? No, we did not create a separate Datasheet. This is because we partially relied on external datasets for one of the data identification steps. Yet, the sources we utilized and the specific components incorporated from each are comprehensively documented in the Data Collection section.

6. Additionally, if you used crowdsourcing or conducted research with human subjects, **without compromising anonymity**...

(a) Did you include the full text of instructions given to participants and screenshots? Yes, they are included in the supplemental material.

(b) Did you describe any potential participant risks, with mentions of Institutional Review Board (IRB) approvals? Yes

(c) Did you include the estimated hourly wage paid to participants and the total amount spent on participant compensation? Yes, they are included in the supplemental material.

(d) Did you discuss how data is stored, shared, and deidentified? Yes

# Supplemental Material

## Instructions Given to Human Participants

We presented the following consent instructions to domain experts and crowdworkers before they began the classification task.

**Introduction and Purpose**

You are being invited to join a research study. The purpose of this study is to identify harmful video content on YouTube. If you agree to join, you will first complete a short questionnaire and later you will be asked to identify harmful YouTube videos. The study will take between 5 and 20 minutes to complete, depending on your performance and videos.

*(This section was shown to crowdworkers only)*

**Compensation**

In this survey, you will first watch 5 YouTube videos. Your task is to classify each video as either "harmful" or "harmless" based on the coding instruction. If you classify more than one out of five videos incorrectly (e.g., saying it was harmless when it was in fact harmful), you will not be eligible to participate in the second part of the study and will not receive compensation as a result. If you classify 4 out of 5 of these videos accurately, you will qualify for the second part of the study, where you will review 25 additional videos and receive $2.

*(This section was shown to crowdworkers only)*

**Potential Risks**

Potential risks related to this research include feeling targeted or hurt by viewing potentially harmful YouTube videos and recalling negative experiences in the past regarding your personal experience with harmful videos online.

**Rights**

Your participation in this research study is voluntary. You can decide to stop the investigation at any time wit out giving us any reasons for your decision and without any negative consequences. If you decide not to participate in this study or if you withdraw from participating at any time, you will not be penalized.

**Questions**

This research has been reviewed and approved by an Institutional Review Board (“IRB”). Information to help you understand research is on-line at http://www.research.ucdavis.edu/policiescompliance/irb-admin/.You may talk to a IRB staff member at 916 703 9158 or by email: HS-IRBEducation@ucdavis.edu, or 2921 Stockton Blvd, Suite 1400, Room 1429, Sacramento, CA 95817 for any of the following. You can also contact the main investigator if you have any questions or want more information about the study at wjo@ucdavis.edu

## Crowdworker Recruitment

Table 5 presents the demographic information of the 544 crowdworkers we recruited via MTurk.

| Demographic | Cohort | % Respondents |
|---|---|---|
| Gender | Male | 324 (59.56%) |
| | Female | 218 (40.07%) |
| | Prefer not to say | 2 (0.39%) |
| Age | 18-24 | 13 (2.39%) |
| | 25-34 | 230 (42.28%) |
| | 35-44 | 234 (43.01%) |
| | 45-54 | 41 (7.54%) |
| | 55-64 | 21 (4.86%) |
| | 65+ | 5 (0.92%) |
| Race & Ethnicity | White | 192 (35.29%) |
| | Black or African American | 5 (0.92%) |
| | American Indian/Native American | 8 (1.47%) |
| | Asian | 329 (60.48%) |
| | Prefer not to say | 7 (1.29%) |
| Employment status | Working full-time | 437 (80.33%) |
| | Working part-time | 81 (14.89%) |
| | Unemployed | 6 (1.10%) |
| | A homemaker | 7 (1.29%) |
| | Retired | 10 (1.84%) |
| | Other | 3 (0.55%) |
| Education | High school diploma or GED | 9 (1.65%) |
| | Some college, but no degree | 31 (5.70%) |
| | Associate or technical degree | 25 (4.41%) |
| | Bachelor’s degree | 400 (73.53%) |
| | Graduate or professional degree | 79 (14.52%) |
| | Heterosexual | 352 (59.74%) |
| | Homosexual | 8 (12.87%) |
| | Bisexual | 178 (32.72%) |
| | Prefer not to say | 7 (12.87%) |
| Disability | Have a disability | 123 (22.61%) |
| | Have no disability | 363 (66.73%) |
| | Prefer not to say | 43 (7.90%) |
| Political party scale | Very liberal | 130 (23.90%) |
| | Somewhat liberal | 177 (32.54%) |
| | Middle of the road | 86 (15.81%) |
| | Somewhat conservative | 80 (14.71%) |
| | Very conservative | 69 (12.68%) |

Table 5. Demographic Information of Mturk Crowdworkers

## Cost Statement

Recruiting crowdworkers via Amazon Mechanical Turk cost us $6,000 to annotate the initial 20,000 videos, with each worker viewing 25 videos for $2. Each video was assigned to three workers, thereby tripling the annotation cost. Additional expenses included platform fees and premiums for hiring master-level workers.

The cost for GPT API varies depending on the tokens required to process video metadata, such as titles, descriptions, and transcripts. Processing the same set of videos using the three GPT-4-Turbo APIs costs approximately $3,500 in total.